%% file: acl_latex.tex
\documentclass[11pt]{article}

\usepackage[preprint]{acl}

\usepackage{times}
\usepackage{latexsym}

\usepackage[T1]{fontenc}

\usepackage[utf8]{inputenc}

\usepackage{microtype}

\usepackage{inconsolata}

\usepackage{graphicx}
\usepackage{algorithm}      
\usepackage{algorithmic}    
\usepackage{amsmath}
\usepackage{amssymb}
\usepackage{booktabs}
\usepackage{colortbl}
\usepackage{multirow}

\definecolor{opdBlue}{HTML}{EAF2FA}
\definecolor{opdBlueStrong}{HTML}{DCEAF7}
\definecolor{opdGold}{HTML}{F8EBC8}
\definecolor{opdGray}{HTML}{F3F6FA}
\definecolor{ruleNavy}{HTML}{17365D}

\newtheorem{proposition}{Proposition}

\title{Not All Disagreement Is Learnable: \\ Token Teachability in On-Policy Distillation}

\author{%
\textbf{Yuanyi Wang}$^{1}$,
\textbf{Su Lu}$^{1}$,
\textbf{Yanggan Gu}$^{1}$,
\textbf{Pengkai Wang}$^{1}$,
\textbf{Yifan Yang}$^{1}$, \\
\textbf{Zhaoyi Yan}$^{2}$,
\textbf{Congkai Xie}$^{2}$,
\textbf{Jianmin Wu}$^{1}$,
\textbf{Hongxia Yang}$^{1,2,3}$\thanks{Corresponding author.} \thanks{wangyuanyi713@gmail.com, hongxia.yang@polyu.edu.hk} \\
$^{1}$The Hong Kong Polytechnic University, PolyU \\
$^{2}$InfiX.ai \quad
$^{3}$Hong Kong Polytechnic University \\Daya Bay Technology and Innovation Research Institute \\
\\
\textbf{Code:} \textcolor{ruleNavy}{\url{https://github.com/wyy-code/TA-OPD}}
}

\begin{document}
\maketitle
\begin{abstract}
On-policy distillation (OPD) trains a student on its own rollouts with token-level teacher supervision. 
Recent selective OPD methods exploit the non-uniformity of OPD signals by prioritizing high-entropy or high-disagreement tokens. 
We revisit this principle and ask: \textit{which token-level teacher signals are actually learnable?} 
Using a fixed-context diagnostic that measures same-context teacher--student KL reduction, we show that raw KL disagreement is a coarse proxy for learning value. 
It conflates \textbf{learnable disagreement}, where the teacher assigns corrective mass to the student's top-\(K\) candidates, with \textbf{incompatible disagreement}, where the teacher places mass mostly off the student's current support.
We formalize this local compatibility as \textit{token teachability} and show that it better predicts fixed-context improvement than raw KL alone. 
Motivated by this finding, we propose \textbf{Teachability-Aware OPD} (TA-OPD), a lightweight token-position selection method that applies OPD loss to high-teachability positions without reward models or verifiers. 
Across Qwen2.5 and Qwen 3 teacher--student settings, TA-OPD often surpasses full-token OPD with only 5\% retained tokens and improves over entropy- and divergence-based baselines. 
Our results reframe selective OPD as selecting learnable teacher signals rather than merely salient tokens.
\end{abstract}

\section{Introduction}

Knowledge distillation has long been used to transfer the behavior of a large teacher model to a smaller student model \citep{hinton2014distilling,zhou2026model,fang2026knowledge}. 
For large language models (LLMs), a particularly effective variant is on-policy distillation (OPD), which trains a student policy on its own rollouts using token-level supervision from a teacher policy \citep{agarwal2024policy,guminiplm,zhang2025aligndistil}.
Compared with off-policy distillation \citep{zhou2025democratizing,wang2026infigfusion}, OPD supervises the student on states that it actually visits, reducing distribution mismatch between training and deployment. 
However, this also exposes a central challenge: the teacher provides dense supervision at every token, but not every token-level teacher signal is equally useful.

\begin{figure}[t]
\centering
\includegraphics[width=\columnwidth]{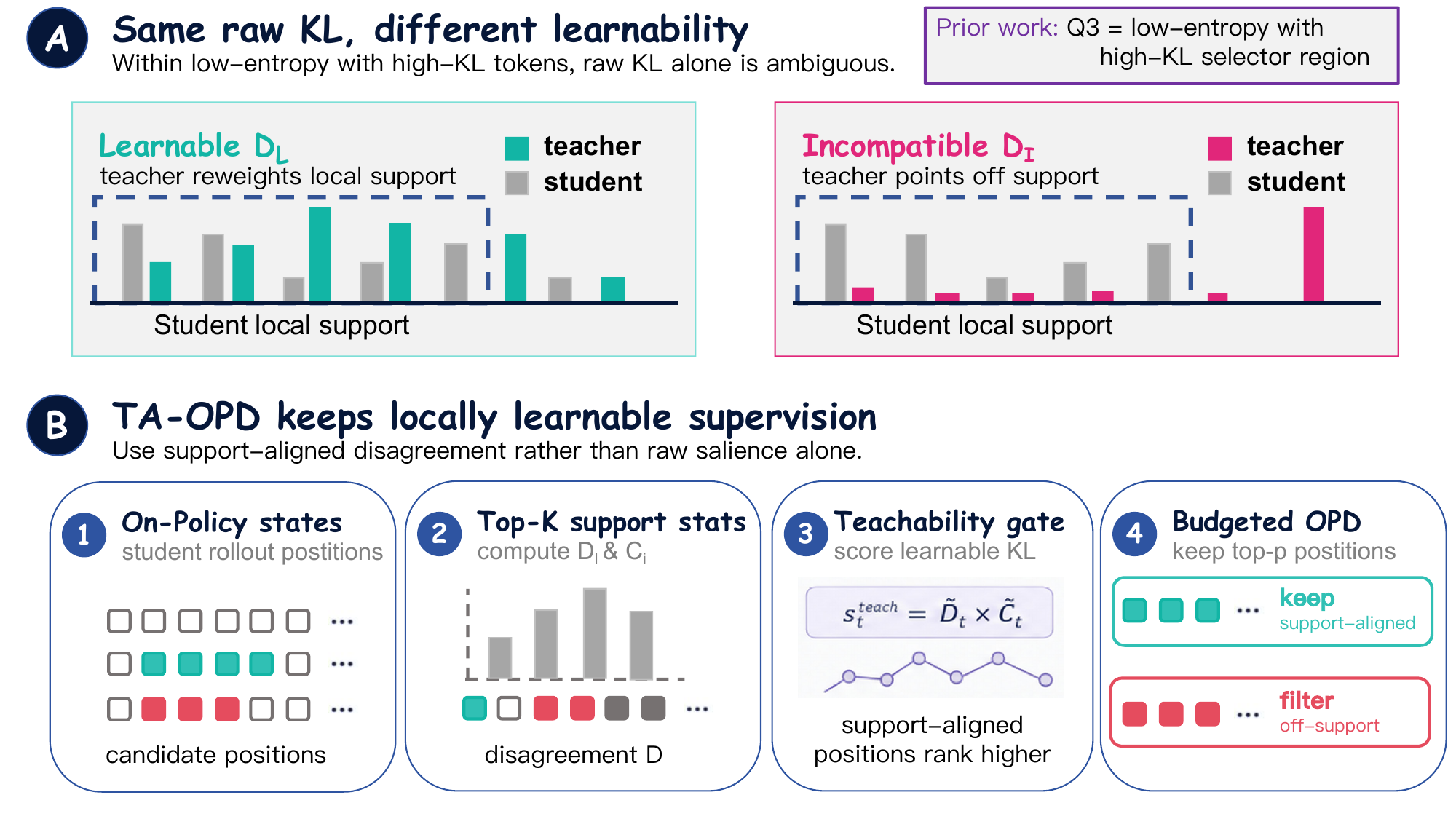}
\caption{\textbf{Token teachability.}
\textbf{A:} Low-entropy, high-KL tokens can contain learnable disagreement \(D^L\), which stays within the student's local support, and incompatible disagreement \(D^I\), which shifts off support.
\textbf{B:} TA-OPD computes disagreement \(D_t\) and compatibility \(C_t\), ranks positions by \(s_t^{\mathrm{teach}}=\tilde{D}_t\tilde{C}_t\), and keeps only high-teachability OPD supervision.}
\label{fig:intro_teachability}
\vspace{-1em}
\end{figure}

Recent work shows that OPD supervision is highly non-uniform \citep{jinentropy,wang2026beyond}. 
Selective OPD therefore allocates token budget to salient positions, such as high-entropy tokens or low-entropy tokens with large teacher--student disagreement \citep{guo2026learning,xie2026llm,xu2026tip}. 
However, these criteria measure uncertainty or disagreement, not the learning effect of the induced token-level supervision. 
In on-policy training, two tokens with similar KL disagreement can lead to very different updates, depending on how the teacher distribution interacts with the student's current predictive state (Fig~\ref{fig:intro_teachability}.A). 
This raises a basic question: \textit{which token-level teacher signals in OPD are actually learnable?}

To answer this question, we need to separate token-level learning value from the rollout in which a token appears. 
End-to-end OPD performance entangles token supervision with sampling noise, downstream context shifts, and interactions across positions. 
We therefore introduce a \emph{fixed-context diagnostic}: we collect student-generated prefixes, freeze them as a context bank, and rescore both the initial and trained students against the same teacher distribution. 
For each token, we measure the same-context reduction in teacher--student KL and ask which pre-update disagreement signals predict this local improvement. 
This diagnostic allows us to study whether a token-level teacher signal is aligned with useful learning, rather than merely salient under entropy or raw KL.

Our diagnostic shows that high KL disagreement is not a uniform learning signal. 
Even in the low-entropy, high-divergence regime targeted by prior selectors, raw disagreement mixes two qualitatively different cases: 
In \textbf{learnable disagreement}, the teacher assigns corrective mass to the student's top-\(K\) candidates, yielding high student-support coverage and better alignment with KL-disagreement reduction. 
In \textbf{incompatible disagreement}, teacher mass falls mostly outside the student's current support, producing a large KL gap but weak local improvement. 
We refer to this local compatibility between teacher correction and student predictive support as \emph{token teachability}. 
Fig~\ref{fig:intro_teachability}.A illustrates this distinction.

Motivated by this diagnostic, we reformulate selective OPD as selecting learnable supervision rather than merely salient tokens. 
We instantiate this principle as \textbf{Teachability-Aware OPD} (TA-OPD), which scores each response position by support-aligned teacher--student disagreement and applies the OPD loss only to high-teachability positions. 
TA-OPD uses teacher and student token probabilities available during OPD, optionally approximated with a lightweight top-\(K\) support, and requires no reward model or verifier (Fig~\ref{fig:intro_teachability}.B).

Across Qwen3 and Qwen2.5 teacher--student settings varying in scale, reasoning ability, and backbone, TA-OPD shows that token quality can outweigh token count. 
With only 5\% retained tokens, TA-OPD often matches or surpasses full-token OPD and improves over budget-matched entropy- and divergence-based methods. 
These results support our central claim: \textit{OPD gains depend on teachable teacher signals, not merely dense supervision.}

Our contributions are summarized as follows. 
\textbf{(i)} we introduce a fixed-context diagnostic that measures same-context KL disagreement and links pre-update token signals to local OPD improvement. 
\textbf{(ii)} using this diagnostic, we formalize \emph{token teachability} and show that raw KL disagreement conflates \textbf{learnable} and \textbf{incompatible} disagreement; 
only support-aligned disagreement reliably predicts useful learning. 
\textbf{(iii)} we propose \textbf{Teachability-Aware OPD} (TA-OPD), a lightweight token-position selection method that applies OPD loss to high-teachability positions without reward models or verifiers. 
\textbf{(iv)} we validate both the finding and the method across Qwen3 and Qwen2.5 settings, showing signficant performance with only 5\% retained tokens.

\section{Related Work}

\textbf{LLM distillation.}
Knowledge distillation transfers a stronger teacher model's behavior to a smaller student by matching outputs, intermediate representations, or generated trajectories \citep{hinton2014distilling,zhou2026model,fang2026knowledge}. 
For LLMs, distillation compresses reasoning, instruction-following, and alignment capabilities under limited compute \citep{zhou2025democratizing,wang2026infigfusion}. 
Off-policy distillation trains on teacher- or human-generated trajectories, which may diverge from student-visited states, while on-policy distillation supervises the student on its own rollouts with token-level teacher distributions \citep{agarwal2024policy,guminiplm,zhang2025aligndistil}. 
We focus on this OPD setting to examine which teacher signals are truly learnable.
\\
\textbf{Selective OPD.}
OPD provides dense token-level supervision, but recent work shows that useful signal is highly non-uniform \citep{yang2026learning}. 
Entropy-based criteria prioritize uncertain tokens \citep{jinentropy,guo2026learning}, and divergence-based criteria highlight positions where the teacher strongly disagrees with the student \citep{xie2026llm,wang2026beyond}. 
TIP \citep{xu2026tip} formalizes this with a two-axis taxonomy of student entropy and teacher--student divergence, enabling token-efficient training. 
Our work complements this line by assessing whether informative tokens are also learnable, showing that high divergence can mix support-aligned and off-support supervision.
\\
\textbf{Teacher--student compatibility.}
Distillation quality depends on both teacher strength and compatibility with the student policy. 
Recent OPD analyses \citep{li2026rethinking} suggest gains arise from alignment on high-probability student tokens and compatible reasoning patterns. 
Self-distillation \citep{zhao2026self,agarwal2024policy} and context-conditioned distillation \citep{ye2026policy,zhang2026opsdl} further emphasize that effective supervision depends on information state and trajectory distribution. 
We operationalize compatibility at the token level by decomposing disagreement into learnable and incompatible components, reframing selective OPD as identifying signals that can be absorbed by the student.

\section{Diagnosing Token Teachability}
\label{sec:token_teachability}

This section diagnoses \textit{which OPD tokens are actually learnable}. 
Using TIP's entropy--KL plane as a reference, we focus on Q3, the low-entropy with high-KL region of confident teacher--student disagreement \citep{xu2026tip}. 
We show that Q3 is heterogeneous: raw KL mixes support-aligned corrections with off-support mismatch. 
We first define fixed-context token gain, then decompose disagreement by local support.

\begin{figure*}[t]
\centering
\includegraphics[width=\textwidth]{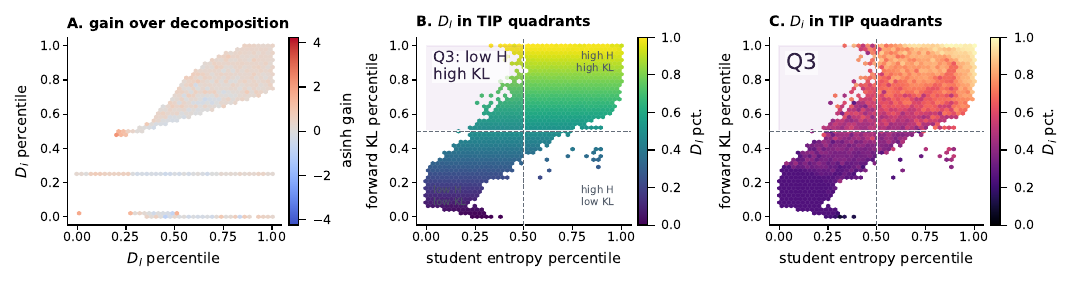}
\caption{\textbf{Local-support decomposition.} 
A: fixed-context gain over learnable disagreement \(D^L\) and incompatible disagreement \(D^I\). 
B--C: \(D^L\) and \(D^I\) projected onto TIP's entropy--KL plane; Q3 denotes the low-entropy/high-KL region. 
Teachability separates support-aligned corrections from off-support mismatch within Q3.}
\label{fig:local_support_phase}
\end{figure*}

\subsection{Fixed-Context Token Gain}
\label{sec:fixed_context_gain}

OPD trains on student-visited contexts \(c_t=(x,y_{<t})\), the prefix generated by the current student. 
Raw KL at \(c_t\) does not by itself reveal learning value, since end-to-end performance also depends on rollout noise and downstream context changes. 
We therefore freeze a bank of on-policy prefixes and rescore checkpoints on the same states.

For \(p_\theta^{i,t}=p_\theta(\cdot\mid c_{i,t})\), define the empirical token gain
\begin{equation}
G_{i,t}^{\mathrm{fix}}
=
D_{\mathrm{KL}}\!\left(p_T^{i,t}\,\|\,p_{\theta_0}^{i,t}\right)
-
D_{\mathrm{KL}}\!\left(p_T^{i,t}\,\|\,p_{\theta_\tau}^{i,t}\right).
\label{eq:fixed_context_gain}
\end{equation}
Positive \(G_{i,t}^{\mathrm{fix}}\) means the trained student is closer to the teacher on the same prefix. 
This diagnostic controls for rollout resampling variance and measures local KL reduction rather than answer-level success.
\begin{proposition}[Gradient alignment]
\label{prop:token_gain}
Let \(\mathcal L_{\mathrm{fix}}\) be differentiable and \(\beta\)-smooth. Let \(\ell_t\) be a token OPD loss. For the update induced by \(\ell_t\), let \(G_t\) be the fixed-context loss reduction. Then
\begin{equation}
\begin{aligned}
G_t
&=
\eta\left\langle
\nabla_\theta\mathcal L_{\mathrm{fix}}(\theta),
\nabla_\theta\ell_t(\theta)
\right\rangle
+R_t,\\
|R_t|&\le
\frac{\beta\eta^2}{2}\|\nabla_\theta\ell_t(\theta)\|_2^2 .
\end{aligned}
\label{eq:token_gain_alignment}
\end{equation}
\end{proposition}

Thus, useful tokens are not merely high-KL tokens; 
their gradients must align with fixed-context improvement.  
Appendices~\ref{app:fixed_context_protocol} and~\ref{app:token_gain_proof} give the diagnostic protocol and proof.

\subsection{Local-Support Decomposition}
\label{sec:local_support_decomp}

We now ask why high KL can fail to be learnable.  At context \(c_t\), define the student's \emph{local support} as its top-\(K\) token set \(S_t^S(K)\); these are the candidates the student currently considers plausible.  Let \(S_t^T(K)\) be the teacher top-\(K\) set and \(U_t=S_t^S(K)\cup S_t^T(K)\).  We measure local disagreement on \(U_t\):
\begin{equation}
D_t
=
D_{\mathrm{KL}}\!\left(\bar p_T^{U_t}\,\|\,\bar p_\theta^{U_t}\right),
\label{eq:sec3_local_disagreement}
\end{equation}
where \(\bar p^{U_t}\) is \(p(\cdot\mid c_t)\) renormalized on \(U_t\).  To measure whether the teacher correction is reachable, define compatibility mass
\begin{equation}
C_t=\sum_{v\in S_t^S(K)}p_T(v\mid c_t).
\label{eq:sec3_compatibility_mass}
\end{equation}
High \(C_t\) means the teacher mostly reweights tokens already in the student's local support; low \(C_t\) means the teacher points off-support.
After robust normalization to \([0,1]\), we decompose disagreement as
\begin{equation}
D_t^L=\widetilde D_t\widetilde C_t,\qquad
D_t^I=\widetilde D_t(1-\widetilde C_t).
\label{eq:sec3_decomposed_disagreement}
\end{equation}
\(D_t^L\) is \emph{learnable disagreement}: the correction is large and locally reachable. 
\(D_t^I\) is \emph{incompatible disagreement}: the correction is large but off-support. 
Fig~\ref{fig:local_support_phase}A links this split to fixed-context gain, and Fig~\ref{fig:local_support_phase}B--C show that Q3 region mixes both signals. 
Thus, teachability measures whether disagreement is locally absorbable, rather than whether the token is uncertain or high-KL. 
Appendix~\ref{app:support_decomposition} gives details.

\begin{figure*}[t]
\centering
\includegraphics[width=\textwidth]{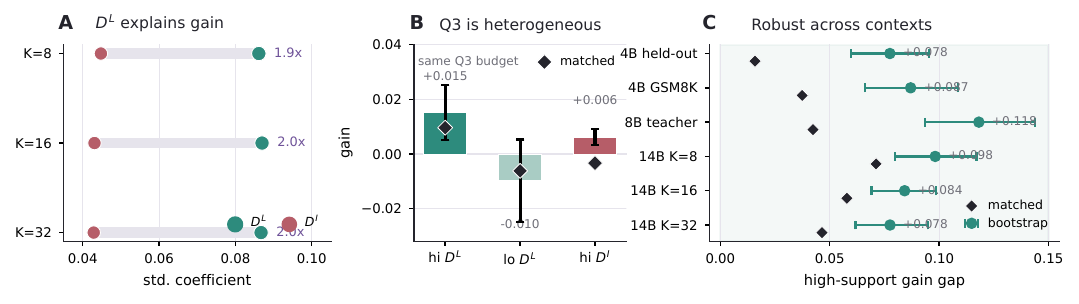}
\caption{\textbf{Fixed-context evidence for token teachability.} 
A: \(D^L\) has about twice the standardized coefficient of \(D^I\). 
B: within Q3, high-\(D^L\) tokens are beneficial, while low-\(D^L\) and high-\(D^I\) tokens are weak or harmful. 
C: high-support gain gaps remain positive across held-out contexts, GSM8K-COT, larger teachers, and support sizes.}
\label{fig:sec3_teachability_story}
\end{figure*}

\subsection{Learnable Disagreement Predicts Fixed-Context Gain}
\label{sec:learnable_explains_gain}

We next test whether this decomposition predicts \(G_{i,t}^{\mathrm{fix}}\). 
For each \(K\), we fit standardized token-level regressions with student entropy, local disagreement, token position, and teacher entropy as controls, and report prompt-cluster bootstrap intervals.

Fig~\ref{fig:sec3_teachability_story}A summarizes the regression evidence. 
Across \(K=8,16,32\), \(D^L\) has roughly twice the coefficient of \(D^I\) (\(0.086\)--\(0.087\) vs. \(0.043\)--\(0.045\)), with positive bootstrap gaps (\(+0.041\)--\(+0.044\)). 
Thus, raw KL is a coarse proxy: it mixes useful corrections with off-support mismatch, while learnable disagreement carries the stable gain signal.
Full analysis are reported in Appendix~\ref{app:regression_diagnostics}.

\subsection{The Low-Entropy with High-Divergence Region is Not Uniformly Teachable}
\label{sec:q3_heterogeneous}

Following TIP, we use Q3 to denote the low-entropy with high-KL quadrant, where the student is confident yet disagrees with the teacher. 
This region is a natural target for selective OPD, but our decomposition asks whether its disagreement is uniformly teachable.

\begin{figure*}[t]
\centering
\includegraphics[width=\textwidth]{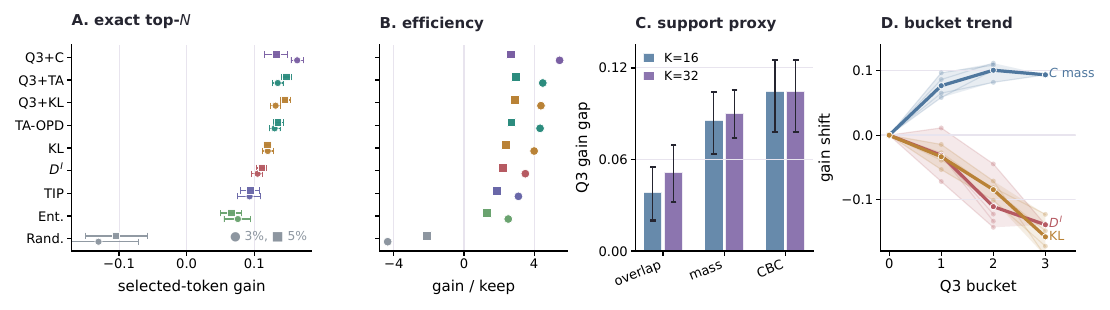}
\caption{\textbf{Q3 controls.} 
A--B: exact top-\(N\) comparisons under matched token counts; B reports gain per kept token. 
C: support proxies yield positive high--low gain gaps inside Q3 at \(K=16,32\). 
D: bucket trends (thin: individual diagnostics; bold: mean) show support mass tracks gain, whereas raw KL and \(D^I\) decline.}
\label{fig:q3_controls}
\end{figure*}

Fig~\ref{fig:sec3_teachability_story}B shows that it is not. 
In a Q3-restricted intervention with matched budgets, high-\(D^L\) tokens yield positive gain, whereas low-\(D^L\) tokens are negative and high-\(D^I\) tokens are weak. 
Thus, Q3 is informative but coarse: teachability separates learnable high-KL supervision from off-support mismatch.
Fig~\ref{fig:q3_controls} controls for token count. 
Q3+TA outperforms TIP, entropy, and random selection under exact top-\(N\) matching (A--B), the effect holds across support proxies (C), and support mass tracks gain better than raw KL or \(D^I\) across buckets (D). 
Bucket trends further show that support mass tracks gain, while raw KL and incompatible disagreement decline.
These controls indicate that Q3 needs a teachability filter, not merely a larger token budget.

\subsection{Robustness Across Contexts and Teachers}
\label{sec:teachability_robustness}

Finally, we test whether the separation depends on a single prompt shard or teacher. 
We repeat the fixed-context diagnostic on held-out math prompts, GSM8K-COT prompts \citep{cobbe2021training,eval-harness}, and stronger 8B/14B teachers. 
Fig~\ref{fig:sec3_teachability_story}C and Table~\ref{tab:scale-context-robustness} show positive high-support gaps across contexts, teachers, and support sizes, with all main prompt-cluster intervals above zero. 
This supports our claim that teachability is a local property of OPD supervision rather than an artifact of one run. 
Appendix~\ref{app:q3_heterogeneity} reports the full exact-budget, bucket, and proxy checks.

\input{tables/tab_scale_context_robustness.tex}

\section{Teachability-Aware OPD}
\label{sec:ta_opd}

The fixed-context diagnostic in Section~\ref{sec:token_teachability} suggests selecting token positions by learnable disagreement rather than raw salience. 
We therefore use teachability as the main selection signal. 
Student entropy is kept separate, serving only as a baseline axis and an optional mixture to test complementarity.

\subsection{Budgeted OPD Objective}
\label{sec:budgeted_opd}

For a rollout batch, let \(\mathcal I\) be all valid response positions after padding and trainer masks. 
At position \(t\), let \(c_t=(x,y_{<t})\) denote the on-policy context, and let \(p_\theta(\cdot\mid c_t)\) and \(p_T(\cdot\mid c_t)\) be the student and teacher next-token distributions over vocabulary \(V\). 
Full reverse-KL OPD uses
\begin{equation}
\begin{aligned}
&\ell_t^{\mathrm{OPD}}
=
D_{\mathrm{KL}}\!\left(p_\theta(\cdot\mid c_t)\,\|\,p_T(\cdot\mid c_t)\right)
\\
&=
\sum_{v\in V}p_\theta(v\mid c_t)
\big[\log p_\theta(v\mid c_t)-\log p_T(v\mid c_t)\big].
\end{aligned}
\label{eq:opd_token_loss}
\end{equation}
Sampled-token OPD may instead use 
\(\widehat\ell_t^{\mathrm{OPD}}=\log p_\theta(y_t\mid c_t)-\log p_T(y_t\mid c_t)\); our selector is agnostic to this estimator. 
Given a binary mask \(m_t\in\{0,1\}\), budgeted OPD optimizes
\begin{equation}
\mathcal L_{m}(\theta)
=
\frac{1}{\sum_{t\in\mathcal I}m_t}
\sum_{t\in\mathcal I}m_t\,\ell_t^{\mathrm{OPD}}(\theta),
\label{eq:masked_opd_loss}
\end{equation}
with \(\widehat\ell_t^{\mathrm{OPD}}\) substituted when used by the trainer. 
This reverse-KL loss is the training objective; the fixed-context KL reduction in Section~\ref{sec:token_teachability} is used only for diagnosis. 
Token budgets refer to KL-supervised response positions, not proportional wall-clock savings.

\subsection{Teachability Score}
\label{sec:teachability_score}

For each position, define the student and teacher top-\(K\) sets
\begin{equation}
\begin{aligned}
S_t^S(K)=\operatorname{TopK}\big(p_\theta(\cdot\mid c_t),K\big),
\\
S_t^T(K)=\operatorname{TopK}\big(p_T(\cdot\mid c_t),K\big),
\end{aligned}
\end{equation}
and let \(U_t=S_t^S(K)\cup S_t^T(K)\). 
We measure local teacher--student disagreement on \(U_t\):
\begin{equation}
\begin{aligned}
D_t &=
D_{\mathrm{KL}}\!\left(\bar p_T^{U_t}\,\|\,\bar p_\theta^{U_t}\right),
\\
\bar p^{U_t}(v\mid c_t) &=
\frac{p(v\mid c_t)}{\sum_{u\in U_t}p(u\mid c_t)},\; v\in U_t .
\end{aligned}
\label{eq:local_disagreement}
\end{equation}
The forward direction emphasizes teacher-preferred candidates underweighted by the student. 
We define \emph{student-support coverage} as the teacher probability mass on the student's top-\(K\) support:
\begin{equation}
C_t=
\sum_{v\in S_t^S(K)}p_T(v\mid c_t).
\label{eq:compatibility_mass}
\end{equation}
High \(C_t\) indicates support-aligned correction; low \(C_t\) indicates off-support disagreement.

We put all token-level scores on a common scale using batch-wise robust normalization. 
Let \(\mathcal I_{\mathcal B}\) be the valid response positions in rollout batch \(\mathcal B\). 
For a raw score \(z_t\) at position \(t\), such as \(D_t\) or \(C_t\), let 
\(z_{\mathcal B}=\{z_j:j\in\mathcal I_{\mathcal B}\}\). 
We define
\begin{equation}
\begin{aligned}
&\operatorname{Norm}_{\mathcal B}(z_t)=
\\
&\operatorname{clip}\!\left(
\frac{z_t-Q_{0.05}(z_{\mathcal B})}
{Q_{0.95}(z_{\mathcal B})-Q_{0.05}(z_{\mathcal B})+\epsilon},0,1\right),
\end{aligned}
\label{eq:robust_norm}
\end{equation}
where \(Q_q\) is the \(q\)-quantile and \(\epsilon>0\) prevents division by zero.
We apply this operator to disagreement and compatibility:
\[
\widetilde D_t=\operatorname{Norm}_{\mathcal B}(D_t),
\qquad
\widetilde C_t=\operatorname{Norm}_{\mathcal B}(C_t).
\]
The teachability score is
\begin{equation}
s_t^{\mathrm{teach}}=D_t^L=\widetilde D_t\,\widetilde C_t.
\label{eq:teachability_score}
\end{equation}

The complementary diagnostic term is \(D_t^I=\widetilde D_t(1-\widetilde C_t)\), which captures large but locally incompatible disagreement. 
Thus, teachability captures locally learnable disagreement, whereas entropy captures uncertainty.

\subsection{Token Selection and Baselines}
\label{sec:ta_selection}

Given retention ratio \(\rho\), let \(n=\lceil\rho|\mathcal I|\rceil\). 
TA-OPD keeps the top-\(n\) valid positions by teachability:
\begin{equation}
\begin{aligned}
\mathcal T_\rho^{\mathrm{teach}}
&=
\operatorname{Top}_{n}\big(\{s_t^{\mathrm{teach}}:t\in\mathcal I\}\big),
\\
m_t&=\mathbf 1[t\in\mathcal T_\rho^{\mathrm{teach}}].
\end{aligned}
\label{eq:teachability_mask}
\end{equation}

To isolate teachability from uncertainty and raw disagreement, we compare against budget-matched selectors:
\begin{align}
s_t^{\mathrm{H}} &= \widetilde H_t,
&
s_t^{\mathrm{D}} &= \widetilde D_t, \\
s_t^{\mathrm{TIP}} &= \widetilde H_t+\widetilde D_t-\widetilde H_t\widetilde D_t,
&
s_t^{\mathrm{C}} &= \widetilde C_t,
\label{eq:baselines}
\end{align}
where \(H_t=H(p_\theta(\cdot\mid c_t))\). 
We also report an optional entropy--teachability mixture,
\begin{equation}
s_t^{\mathrm{H+teach}}
=
\widetilde H_t+s_t^{\mathrm{teach}}-\widetilde H_t s_t^{\mathrm{teach}},
\label{eq:entropy_teachability}
\end{equation}
to test complementarity with the TIP entropy axis. 
Full OPD sets \(m_t=1\) for all \(t\in\mathcal I\), and random selection samples \(n\) valid positions uniformly.

In practice, \(D_t\) and \(C_t\) are computed from teacher and student top-\(K\) log-probabilities. 
If teacher scores on \(S_t^S(K)\) are available, Eq.~\ref{eq:compatibility_mass} is exact; 
otherwise we use 
\begin{equation}
\widehat C_t=\sum_{v\in S_t^S(K)\cap S_t^T(K)}p_T(v\mid c_t),
\end{equation}
a lower bound on \(C_t\). 
Unless stated otherwise, \(K=16\). 
TA-OPD requires no reward model, verifier, or additional labels.

\paragraph{Fixed-context selector check.}
Before downstream evaluation, we verify that the diagnostic score can serve as a training mask. 
Table~\ref{tab:selector-intervention} compares budgeted selectors on the same fixed-context bank. 
TA-OPD gives the largest KL reduction and gain per kept token at both 3\% and 5\% budgets, supporting teachability as an OPD allocation rule rather than only a post-hoc diagnostic.

\input{tables/tab_main_selector_intervention.tex}

\input{tables/tab_main_results_qwen3.tex}
\input{tables/tab_budget_sweep_qwen3.tex}

\section{Experiments}
\label{sec:experiments}

\subsection{Setup}
\label{sec:exp_setup}

\textbf{Models.}
We evaluate four teacher--student pairs: 
\\
(1) Qwen3-4B to Qwen3-1.7B
\\
(2) Qwen3-8B-GRPO to Qwen3-4B
\\
(3) Qwen3-14B to Qwen3-4B
\\
(4) DeepSeek-R1-Distill-Qwen-14B \citep{guo2025deepseek} to Qwen2.5-3B \citep{qwen2025qwen25technicalreport}.

\noindent
The Qwen3-8B teacher is GRPO-tuned \citep{shao2024deepseekmath}, 
so the settings vary \textit{scale, capability, and backbone architecture}.
\\
\textbf{Training data and benchmarks.}
Training prompts are sampled from DAPO~\citep{yu2026dapo}.  
We evaluate math, coding, factual QA, and instruction following on AIME24~\citep{aime24}, AIME25~\citep{aime25}, GPQA-Diamond~\citep{reingpqa}, HumanEval~\citep{chen2021evaluating}, IFEval~\citep{zhou2023instruction}, and MATH-500~\citep{hendrycks2measuring},
based on the EvalScope \citep{evalscope_2024}.
Each result reports the mean and standard deviation over five evaluation seeds from the same trained checkpoint.
\\
\textbf{Baselines and budgets.}
We compare full OPD with budgeted token selectors: entropy-only \citep{jinentropy}, TIP-style entropy+divergence \citep{xu2026tip}, Teachability-Aware OPD (TA-OPD), and TA-OPD+Entropy.  Unless otherwise stated, support statistics use \(K=16\), and budgets refer to KL-supervised response-token positions rather than wall-clock savings.
\\
\textbf{Implementation.}
All runs use the same sampled-token OPD pipeline as the fixed-context analysis and are trained on 64 NVIDIA H800 GPUs.

\subsection{Main Results}
\label{sec:exp_main_results}

Table~\ref{tab:main-results-qwen3} compares all methods at a 10\% supervised-token budget. 
Across the four teacher--student settings, TA-OPD obtains the best average score in every group: 44.89 for Qwen3-4B to Qwen3-1.7B, 56.87 for Qwen3-8B-GRPO to Qwen3-4B, 54.65 for Qwen3-14B to Qwen3-4B, and 30.62 for the cross-backbone DeepSeek-R1-Distill-Qwen to Qwen2.5 setting. 
This supports the main claim that low-budget OPD can preserve useful supervision when the retained tokens are teachable.

The gains are strongest when the student is smaller or the teacher--student pair is mismatched. 
For Qwen3-4B to Qwen3-1.7B, TA-OPD improves the average over Full OPD from 42.37 to 44.89 and leads on AIME24, GPQA-Diamond, and IFEval. 
For the cross-backbone setting, TA-OPD improves over both Base and Full OPD in average score, while TA-OPD+Entropy gives the best AIME24, AIME25, and GPQA-Diamond scores. 
These results suggest that teachability is especially useful when dense teacher supervision contains more incompatible or noisy token signals.

For stronger Qwen3 teachers with a 4B student, the results have similar trend. 
With the GRPO-tuned 8B teacher, TA-OPD has the best average and leads on AIME24 and MATH-500, while TIP or entropy remain competitive on HumanEval and IFEval. 
With the 14B teacher, TA-OPD nearly matches Full OPD in average score and leads HumanEval, while TA-OPD+Entropy is strongest on AIME25. 
Thus, teachability is not a universal per-benchmark booster; it is a token-allocation rule that improves the quality of supervised positions under a constrained OPD budget.

\subsection{Budget Sensitivity}
\label{sec:exp_budget_scale_arch}

Table~\ref{tab:budget-sweep-qwen3} sweeps 5\%, 10\%, 30\%, and 50\% budgets for the two Qwen3-4B student settings. 
The results are not monotonic in budget, indicating that more KL-supervised tokens do not always yield better downstream performance.

For Qwen3-8B-GRPO to Qwen3-4B, low budgets are already competitive: TA-OPD+Entropy reaches the best 5\% average score (57.89), while TA-OPD gives the best MATH-500 score at 5\% and the best average among teachability-based selectors at 50\% (57.90). 
TIP performs best at 30\% average, showing that uncertainty and divergence can still help on some benchmark mixtures. 
However, the strong 5\% results indicate that much of the useful OPD signal is concentrated in a small set of teachable positions.

For Qwen3-14B to Qwen3-4B, the best average appears at 10\% with TA-OPD (54.65), while TA-OPD+Entropy is strongest at 5\% (54.47). 
Increasing the budget to 30\% or 50\% does not consistently improve the average, and several metrics regress relative to the 5--10\% settings. 
This is consistent with the fixed-context analysis: OPD is token quality limited rather than purely budget-limited, and the optimal ratio depends on the teacher, student, and benchmark mix.
Appendix~\ref{app:budget_downstream} provides fixed-context budget curves, macro-average budget summaries, and early downstream boundary checks.
\\
\textbf{Remark 1:}
Appendix~\ref{app:budget_downstream} reports the ablation study.

\section{Conclusion}
\label{sec:conclusion}

We studied which token-level teacher signals are actually learnable in on-policy distillation. 
Using a fixed-context diagnostic, we showed that raw KL disagreement is a coarse proxy for learning value: it conflates learnable disagreement, where teacher mass has high coverage over the student's top-\(K\) support, with incompatible disagreement, where teacher mass falls mostly off support. 
We formalized this support-aware learning value as \emph{token teachability} and used it to design Teachability-Aware OPD, a lightweight token-position selection method without reward model or verifier. 
Across Qwen3 and Qwen2.5 teacher--student settings, TA-OPD often surpasses full-token OPD and improves over entropy- and divergence-based selectors. 
These results suggest that selective OPD should prioritize locally learnable teacher signals rather than merely salient tokens, reframing OPD as a compatibility-aware supervision problem.

\newpage

\section{Limitations}
\label{app:limitations}

Our analysis focuses on OPD for math-heavy reasoning prompts and Qwen-family teacher--student pairs, with one cross-backbone distillation setting.  Although the fixed-context diagnostic is model-agnostic, broader coverage over multilingual data, dialogue tasks, code-specialized teachers, and non-Qwen backbones would further test the generality of token teachability.  TA-OPD also selects supervised token positions rather than pruning transformer computation, so the reported token budget should be interpreted as a supervision budget, not a proportional wall-clock speedup.  Finally, our diagnostic measures same-context KL reduction; it is designed to explain local learning signals and should be paired with downstream evaluation when making deployment claims.

\bibliography{custom}

\appendix

\section{Use of AI Assistants.}
We used large language model assistants for language polishing, grammar checking, and formatting refinement during paper preparation. 
All technical ideas, method design, experiments, analyses, claims, and final writing were developed, verified, and approved by the authors. 
The authors take full responsibility for the content of the submission.

\section{Ethical Considerations}
\label{app:ethics}

TA-OPD is a training-time method for filtering teacher supervision.  It can reduce exposure to noisy or incompatible teacher signals, but it does not provide guarantees about factuality, safety, bias, or harmful content in the distilled model.  If the teacher produces unsafe behavior, selective distillation may still transfer parts of that behavior when those tokens appear locally learnable to the student.  Practical use should therefore combine TA-OPD with standard data filtering, safety evaluation, and task-specific deployment checks.  Our experiments use public benchmark-style data and do not rely on private user data.

\section{Fixed-Context Diagnostic Protocol}
\label{app:fixed_context_protocol}

\paragraph{Context bank.}
For each prompt, we sample student rollouts and store valid response prefixes \(c_{i,t}=(x_i,y_{i,<t})\).  All checkpoints are scored on this same bank, so diagnostic differences cannot come from resampling different rollouts.

\paragraph{Support metrics.}
For support size \(K\), \(S_t^S(K)\) and \(S_t^T(K)\) denote the student and teacher top-\(K\) token sets.  We compute disagreement on \(S_t^S(K)\cup S_t^T(K)\), and compatibility \(C_t^{(K)}\) as teacher mass on \(S_t^S(K)\).  Scalars are robustly normalized within a batch by clipping the 5--95 percentile range to \([0,1]\).

\paragraph{Fixed-context gain.}
When full-vocabulary probabilities are available, we compute Eq.~\ref{eq:fixed_context_gain}.  Otherwise, top-\(K\) support diagnostics are used only for robustness checks.  Because token gains are heavy-tailed and correlated within a rollout, all confidence intervals use prompt-cluster bootstrap.

\input{tables/tab_section3_dataset_manifest.tex}

\paragraph{Gradient view.}
Proposition~\ref{prop:token_gain} gives the local optimization view used in Section~\ref{sec:fixed_context_gain}.  The proof is included in Appendix~\ref{app:token_gain_proof}.

\section{Local-Support Decomposition}
\label{app:support_decomposition}

\paragraph{Top-\(K\) supports.}
For each fixed context \(c_t\), we store the student and teacher top-\(K\) sets \(S_t^S(K)\) and \(S_t^T(K)\).  The union \(U_t=S_t^S(K)\cup S_t^T(K)\) is used for local disagreement.  For \(v\in U_t\), the restricted distribution is
\[
\bar p^{U_t}(v\mid c_t)=
\frac{p(v\mid c_t)}{\sum_{u\in U_t}p(u\mid c_t)}.
\]
This keeps the diagnostic focused on locally salient alternatives rather than full-vocabulary tail noise.

\paragraph{Normalization.}
All scalar token scores are normalized within a rollout batch:
\[
\widetilde z_t=
\operatorname{clip}\!\left(
\frac{z_t-Q_{0.05}(z_{\mathcal B})}
{Q_{0.95}(z_{\mathcal B})-Q_{0.05}(z_{\mathcal B})+\epsilon},
0,1\right),
\]
where \(Q_q\) is the \(q\)-quantile.  We use the same normalization for entropy, disagreement, and compatibility-derived scores.

\paragraph{Support proxies.}
The main compatibility mass is \(C_t=\sum_{v\in S_t^S(K)}p_T(v\mid c_t)\).  We also audit simpler proxies: top-\(K\) overlap fraction, top-\(K\) Jaccard, shared teacher top-\(K\) mass, and a contrastive binary compatibility score (CBC).  These proxies ask the same question: how much of the teacher correction remains near the student's local support?

\input{tables/tab_support_proxy_audit.tex}

\section{Low-Entropy with High-Divergence Heterogeneity Diagnostics}
\label{app:q3_heterogeneity}

\paragraph{Live quadrant intervention.}
We train three masks restricted to the low-entropy with high-divergence (Q3) region: high \(D^L\), low \(D^L\), and high \(D^I\).  All keep roughly the same token ratio.  The high-\(D^L\) mask is the only clearly positive intervention; low-\(D^L\) is negative, and high-\(D^I\) is weak.

\input{tables/tab_within_q3_live.tex}

\begin{figure*}[t]
\centering
\includegraphics[width=\textwidth]{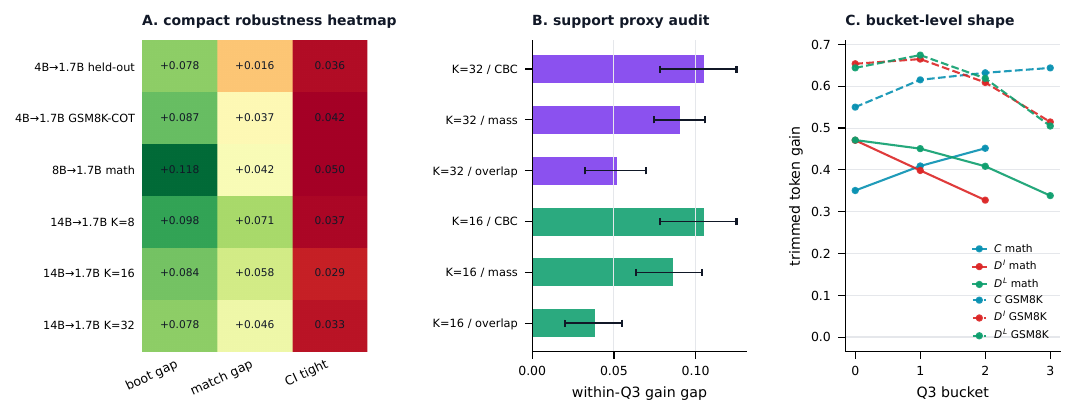}
\caption{Additional low-entropy with high-divergence and robustness evidence.  The panels visualize the support-proxy audit, compact robustness statistics, and bucket-level gain shapes used by Sections~\ref{sec:q3_heterogeneous}--\ref{sec:learnable_explains_gain}.}
\label{fig:app_section3_p1_evidence}
\end{figure*}

\paragraph{Exact-budget fixed-context controls.}
Fig~\ref{fig:q3_controls} summarizes the main Q3 control evidence.  Here we report the full exact top-\(N\) table behind Panels A--B.  Every selector keeps the same number of target tokens; support-aligned selectors remain positive, while random selection is consistently negative.
Appendix Fig~\ref{fig:app_section3_p1_evidence}C shows the bucket-level gain shapes used as a nonlinear sanity check.

\input{tables/tab_matched_topn.tex}

\paragraph{Support-definition audit.}
The quadrant result is stable across support proxies.  Teacher mass on student support, top-\(K\) overlap, Jaccard overlap, and CBC all produce positive high--low support gaps inside the low-entropy with high-divergence region.
Appendix Fig~\ref{fig:app_section3_p1_evidence}A gives a compact heatmap of the robustness gaps in Table~\ref{tab:scale-context-robustness}.
Appendix Fig~\ref{fig:app_section3_p1_evidence}B visualizes the support-proxy audit.

\section{Regression Diagnostics for Learnable Disagreement}
\label{app:regression_diagnostics}

\paragraph{Specification.}
For each token, the dependent variable is the fixed-context gain \(G_{i,t}^{\mathrm{fix}}\) in Eq.~\ref{eq:fixed_context_gain}.  All scalar predictors are standardized within the diagnostic dataset.  The baseline includes student entropy, local disagreement, their interaction, normalized token position, and teacher entropy.  We then compare two extensions: adding compatibility \(C_t\) alone, or replacing raw disagreement with the decomposed pair \(D_t^L,D_t^I\).  Confidence intervals use prompt-cluster bootstrap.

\paragraph{Why coefficients, not only \(R^2\).}
Token gains are heavy-tailed and noisy because a single local update can affect later prefixes and formatting tokens.  We therefore use standardized coefficients and prompt-cluster gaps as the main diagnostic; \(R^2\) is reported only as an incremental sanity check.

\input{tables/tab_decomposition_regression.tex}

\paragraph{Nonlinear sanity check.}
We also bin residualized gains by \(D^L\), \(D^I\), and \(C_t\) using the same fixed-context bank.  The binned curves do not indicate that the separation in Fig~\ref{fig:sec3_teachability_story}A is an artifact of a purely linear model.  The source artifacts are archived as \texttt{standardized\_regression\_coefficients.csv} and \texttt{spline\_sanity\_bins.csv} under \texttt{figs/}.

\section{Budget and Downstream Boundary Checks}
\label{app:budget_downstream}

\paragraph{Budget sweep.}
We sweep the effective KL-supervised token budget for TA-OPD and low-entropy with high-divergence compatibility controls.  The fixed-context gain peaks around 3--5\% and saturates at 10\%, suggesting that selected supervision is quality-limited rather than simply budget-limited.

\input{tables/tab_budget_curve.tex}

\paragraph{Macro-average budget view.}
Table~\ref{tab:budget-ratio-qwen3} compresses the Qwen3-4B student budget sweep from Table~\ref{tab:budget-sweep-qwen3} into benchmark macro-averages.  The trend is not monotonic: TA-OPD or TA-OPD$_{+Ent.}$ is strongest at several low-budget points, while larger budgets can help or hurt depending on the teacher and benchmark mix.

\input{tables/tab_budget_ratio_qwen3.tex}

\paragraph{Downstream selector ablation.}
Table~\ref{tab:downstream-selector-ablation} compares alternative token selectors in the Qwen3-8B-GRPO to Qwen3-4B setting at the same 10\% supervised-token budget.  TA-OPD obtains the best average score (54.65), ahead of raw KL (53.76), \(C\)-only selection (54.19), Q3-only selection (52.46), and Q3+high-\(C\) selection (53.32).  This supports the main interpretation: useful OPD targets require both a meaningful teacher correction and local support alignment.

\input{tables/tab_downstream_selector_ablation.tex}

\paragraph{Early downstream smoke checks.}
Before the full benchmark suite in Section~\ref{sec:exp_main_results}, we ran small downstream smoke checks on GSM8K-COT, MATH-hard, and capped AIME.  GSM8K-COT followed the fixed-context trend, while MATH-hard and capped AIME were low-resolution boundary checks; we therefore report the complete downstream evidence only in the main benchmark tables and omit the sparse smoke table.

\input{tables/tab_math_category_deltas.tex}

\section{Impact to Model Fusion}
Model fusion aims to aggregate capabilities from multiple LLMs under limited compute, either by parameter-space merging or by distillation-based fusion. 
Recent merging studies characterize scaling behavior and budget-aware parameter management \citep{wang2025model,wang2026mergepipe}, while related work studies post-merge calibration, quantization, and geometry conflict in weight space \citep{wang2026geometry,gu2026featcal,wang2026pmq,wang2026discovering}. 
Another line performs implicit or logit-level fusion through preference optimization and distillation, treating model outputs as transferable supervision signals \citep{gu2026infifpo,wang2026infigfusion}. 
Our work is complementary to both views. 
Rather than asking how to combine models globally, TA-OPD studies which token-level teacher signals are locally absorbable by the current student policy. 
This suggests a compatibility-aware perspective on distillation-based fusion: effective fusion should not only aggregate stronger sources, but also identify the subset of teacher signals that the target model can learn from.

\section{Proof of Proposition~\ref{prop:token_gain}}
\label{app:token_gain_proof}

Let \(g_t=\nabla_\theta \ell_t(\theta)\) and \(\Delta_t=-\eta g_t\). 
By \(\beta\)-smoothness of \(\mathcal{L}_{\mathrm{fix}}\),
\[
\mathcal{L}_{\mathrm{fix}}(\theta+\Delta_t)
=
\mathcal{L}_{\mathrm{fix}}(\theta)
+
\left\langle
\nabla_\theta \mathcal{L}_{\mathrm{fix}}(\theta),
\Delta_t
\right\rangle
+
r_t,
\]
with
\[
|r_t|
\le
\frac{\beta}{2}\|\Delta_t\|_2^2 .
\]
Substituting \(\Delta_t=-\eta g_t\) gives
\[
\mathcal{L}_{\mathrm{fix}}(\theta_t^+)
=
\mathcal{L}_{\mathrm{fix}}(\theta)
-
\eta
\left\langle
\nabla_\theta \mathcal{L}_{\mathrm{fix}}(\theta),
g_t
\right\rangle
+
r_t .
\]
Rearranging,
\[
G_t
=
\eta
\left\langle
\nabla_\theta \mathcal{L}_{\mathrm{fix}}(\theta),
g_t
\right\rangle
-
r_t .
\]
Setting \(R_t=-r_t\) yields the result.

\end{document}

%% file: tables/tab_scale_context_robustness.tex
\begin{table*}[t]
\centering
\small
\setlength{\tabcolsep}{4.0pt}
\renewcommand{\arraystretch}{1.06}
\caption{\textbf{Context and teacher-scale robustness.} 
The gap compares high- vs. low-support tokens inside Q3, the low-entropy/high-KL diagnostic region. 
Bootstrap intervals are prompt-clustered; matched gap denotes the prompt-matched estimate. 
All main 300-context diagnostics have positive intervals.}
\label{tab:scale-context-robustness}
\resizebox{\textwidth}{!}{%
\begin{tabular}{llrrrrr}
\toprule
Setting & Context / support & Contexts & Tokens & Bootstrap gap & 95\% interval & Matched gap \\
\midrule
Qwen3-4B$\rightarrow$1.7B & held-out math, $K=16$ & 300 & 57,600 & +0.078 & [+0.060, +0.095] & +0.016 \\
Qwen3-4B$\rightarrow$1.7B & GSM8K-COT, $K=16$ & 300 & 57,600 & +0.087 & [+0.066, +0.109] & +0.037 \\
Qwen3-8B$\rightarrow$1.7B & held-out math, $K=16$ & 300 & 57,600 & +0.118 & [+0.094, +0.144] & +0.042 \\
Qwen3-14B$\rightarrow$1.7B & held-out math, $K=8$ & 300 & 57,600 & +0.098 & [+0.080, +0.117] & +0.071 \\
\rowcolor{opdBlue}
Qwen3-14B$\rightarrow$1.7B & held-out math, $K=16$ & 300 & 57,600 & +0.084 & [+0.069, +0.098] & +0.058 \\
Qwen3-14B$\rightarrow$1.7B & held-out math, $K=32$ & 300 & 57,600 & +0.078 & [+0.062, +0.095] & +0.046 \\
\bottomrule
\end{tabular}%
}
\end{table*}

%% file: tables/tab_main_selector_intervention.tex
\begin{table}[t]
\centering
\scriptsize
\setlength{\tabcolsep}{2.1pt}
\renewcommand{\arraystretch}{1.06}
\caption{\textbf{Fixed-context selector check.} 
\(G_{3\%}\) and \(G_{5\%}\) are same-context KL reductions at nominal budgets; \(G/K\) normalizes by actual retained ratio (Keep). 
TA-OPD gives the largest reduction and efficiency. 
\(C\) is teacher mass on student support, and Q3 is TIP's low-entropy/high-divergence region.}
label{tab:selector-intervention}
\label{tab:selector-intervention}
\resizebox{\columnwidth}{!}{%
\begin{tabular}{lrrrrrr}
\toprule
Selector & Keep & $G_{3\%}$ & $G/K_{3\%}$ & $G_{5\%}$ & $G/K_{5\%}$ & Q3 frac. \\
\midrule
\rowcolor{opdBlueStrong}
\textbf{TA-OPD} & 3.4 & \textbf{0.047} & \textbf{1.40} & \textbf{0.049} & \textbf{0.95} & 4.4 \\
\rowcolor{opdGold}
Raw KL & 3.4 & 0.033 & 0.97 & 0.036 & 0.69 & 2.7 \\
Incompat. $D^I$ & 3.5 & 0.025 & 0.72 & -- & -- & 1.4 \\
TIP & 3.3 & 0.023 & 0.69 & 0.026 & 0.50 & 1.6 \\
Random & 3.3 & 0.021 & 0.65 & 0.021 & 0.41 & 3.3 \\
TA-OPD$_{+Ent.}$ & 3.3 & 0.021 & 0.63 & 0.016 & 0.31 & 1.0 \\
Entropy & 3.3 & 0.018 & 0.53 & 0.009 & 0.18 & 0.3 \\
\rowcolor{opdGray}
Q3+high-$C$ & 3.2 & 0.016 & 0.51 & 0.009 & 0.23 & 94.6 \\
\rowcolor{opdGray}
Q3+low-$C$ & 3.2 & 0.010 & 0.32 & -- & -- & 94.7 \\
\rowcolor{opdGray}
Q3-only & 3.2 & 0.010 & 0.31 & 0.002 & 0.05 & 91.7 \\
\bottomrule
\end{tabular}
}
\vspace{0.2em}
\end{table}

%% file: tables/tab_main_results_qwen3.tex
\begin{table*}[t]
\centering
\scriptsize
\setlength{\tabcolsep}{2.8pt}
\renewcommand{\arraystretch}{1.09}
\caption{Main benchmark results at a 10\% supervised-token budget. 
We evaluate Qwen3 teacher--student pairs and a cross-backbone DeepSeek-R1-Distill-Qwen-14B to Qwen2.5-3B pair. 
Entries report mean\(\pm\)std over five evaluation seeds; Avg. is the average score over six benchmarks. 
TA-OPD denotes Teachability-Aware OPD, and TA-OPD$_{+Ent.}$ adds entropy.}
\label{tab:main-results-qwen3}
\resizebox{\textwidth}{!}{%
\begin{tabular}{llrrrrrrr}
\toprule
Setting & Method & Avg. & AIME24 & AIME25 & GPQA-D. & HumanEval & IFEval & MATH-500 \\
\midrule
\multirow{6}{*}{\cellcolor{white}\shortstack[l]{Qwen3-4B\\$\rightarrow$\\Qwen3-1.7B}} & \cellcolor{opdGray}Base & \cellcolor{opdGray}40.77 & \cellcolor{opdGray}10.00$\pm$3.33 & \cellcolor{opdGray}10.00$\pm$3.33 & \cellcolor{opdGray}31.56$\pm$0.76 & \cellcolor{opdGray}60.97$\pm$1.83 & \cellcolor{opdGray}59.52$\pm$1.29 & \cellcolor{opdGray}72.60$\pm$1.00 \\
 & \cellcolor{opdGold}Pure OPD & \cellcolor{opdGold}42.37 & \cellcolor{opdGold}11.67$\pm$5.00 & \cellcolor{opdGold}\textbf{15.00$\pm$1.67} & \cellcolor{opdGold}30.55$\pm$0.25 & \cellcolor{opdGold}59.15$\pm$1.22 & \cellcolor{opdGold}62.85$\pm$0.74 & \cellcolor{opdGold}75.00$\pm$1.20 \\
 & Entropy & 41.46 & 13.34$\pm$3.34 & 8.33$\pm$5.00 & 32.58$\pm$1.77 & 59.45$\pm$0.30 & 60.63$\pm$0.01 & 74.40$\pm$1.00 \\
 & TIP & 43.05 & 18.34$\pm$1.66 & 11.67$\pm$5.00 & 29.29$\pm$2.02 & \textbf{62.50$\pm$0.30} & 60.91$\pm$0.27 & \textbf{75.60$\pm$1.00} \\
 & \cellcolor{opdBlueStrong}TA-OPD & \cellcolor{opdBlueStrong}\textbf{44.89} & \cellcolor{opdBlueStrong}\textbf{20.00$\pm$0.01} & \cellcolor{opdBlueStrong}\textbf{15.00$\pm$5.00} & \cellcolor{opdBlueStrong}\textbf{34.59$\pm$2.27} & \cellcolor{opdBlueStrong}60.98$\pm$0.01 & \cellcolor{opdBlueStrong}\textbf{64.05$\pm$1.02} & \cellcolor{opdBlueStrong}74.70$\pm$0.50 \\
 & \cellcolor{opdBlue}TA-OPD$_{+Ent.}$ & \cellcolor{opdBlue}42.32 & \cellcolor{opdBlue}15.00$\pm$1.67 & \cellcolor{opdBlue}8.34$\pm$1.67 & \cellcolor{opdBlue}32.58$\pm$1.77 & \cellcolor{opdBlue}61.59$\pm$0.61 & \cellcolor{opdBlue}61.73$\pm$0.92 & \cellcolor{opdBlue}74.70$\pm$0.10 \\
\midrule
\multirow{6}{*}{\cellcolor{white}\shortstack[l]{Qwen3-8B\\(GRPO)$\rightarrow$\\Qwen3-4B}} & \cellcolor{opdGray}Base & \cellcolor{opdGray}52.98 & \cellcolor{opdGray}21.66$\pm$1.66 & \cellcolor{opdGray}13.33$\pm$0.01 & \cellcolor{opdGray}\textbf{47.97$\pm$0.51} & \cellcolor{opdGray}77.75$\pm$0.30 & \cellcolor{opdGray}73.57$\pm$2.22 & \cellcolor{opdGray}83.60$\pm$0.20 \\
 & \cellcolor{opdGold}Pure OPD & \cellcolor{opdGold}53.71 & \cellcolor{opdGold}20.00$\pm$3.33 & \cellcolor{opdGold}21.67$\pm$1.67 & \cellcolor{opdGold}47.22$\pm$1.26 & \cellcolor{opdGold}78.36$\pm$0.92 & \cellcolor{opdGold}70.33$\pm$1.20 & \cellcolor{opdGold}84.70$\pm$1.30 \\
 & Entropy & 56.69 & 26.33$\pm$3.10 & 23.33$\pm$3.30 & 47.47$\pm$1.01 & 77.65$\pm$1.21 & \textbf{79.85$\pm$0.37} & 85.50$\pm$0.10 \\
 & TIP & 56.81 & 26.33$\pm$3.60 & 23.67$\pm$3.33 & 47.47$\pm$0.49 & \textbf{80.49$\pm$0.61} & 76.71$\pm$1.11 & 86.20$\pm$0.30 \\
 & \cellcolor{opdBlueStrong}TA-OPD & \cellcolor{opdBlueStrong}\textbf{56.87} & \cellcolor{opdBlueStrong}\textbf{30.00$\pm$1.40} & \cellcolor{opdBlueStrong}21.31$\pm$1.32 & \cellcolor{opdBlueStrong}47.47$\pm$0.51 & \cellcolor{opdBlueStrong}79.88$\pm$0.60 & \cellcolor{opdBlueStrong}75.13$\pm$5.30 & \cellcolor{opdBlueStrong}\textbf{87.40$\pm$0.80} \\
 & \cellcolor{opdBlue}TA-OPD$_{+Ent.}$ & \cellcolor{opdBlue}56.52 & \cellcolor{opdBlue}\textbf{30.00$\pm$3.33} & \cellcolor{opdBlue}\textbf{25.00$\pm$1.67} & \cellcolor{opdBlue}45.70$\pm$2.78 & \cellcolor{opdBlue}78.05$\pm$1.22 & \cellcolor{opdBlue}73.38$\pm$2.22 & \cellcolor{opdBlue}87.00$\pm$0.40 \\
\midrule
\multirow{6}{*}{\cellcolor{white}\shortstack[l]{Qwen3-14B\\$\rightarrow$\\Qwen3-4B}} & \cellcolor{opdGray}Base & \cellcolor{opdGray}52.98 & \cellcolor{opdGray}21.66$\pm$1.66 & \cellcolor{opdGray}13.33$\pm$0.01 & \cellcolor{opdGray}47.97$\pm$0.51 & \cellcolor{opdGray}77.75$\pm$0.30 & \cellcolor{opdGray}73.57$\pm$2.22 & \cellcolor{opdGray}83.60$\pm$0.20 \\
 & \cellcolor{opdGold}Pure OPD & \cellcolor{opdGold}54.64 & \cellcolor{opdGold}\textbf{23.34$\pm$3.34} & \cellcolor{opdGold}21.66$\pm$1.66 & \cellcolor{opdGold}42.17$\pm$0.76 & \cellcolor{opdGold}78.05$\pm$0.61 & \cellcolor{opdGold}\textbf{79.11$\pm$0.74} & \cellcolor{opdGold}83.50$\pm$0.10 \\
 & Entropy & 53.39 & 21.67$\pm$5.00 & 15.00$\pm$1.67 & 46.46$\pm$0.01 & 79.27$\pm$0.61 & 73.94$\pm$4.06 & \textbf{84.00$\pm$0.80} \\
 & TIP & 53.62 & 20.00$\pm$0.01 & 16.66$\pm$3.33 & \textbf{47.98$\pm$2.53} & 78.05$\pm$0.61 & 76.62$\pm$1.57 & 82.40$\pm$0.01 \\
 & \cellcolor{opdBlueStrong}TA-OPD & \cellcolor{opdBlueStrong}\textbf{54.65} & \cellcolor{opdBlueStrong}\textbf{23.34$\pm$3.34} & \cellcolor{opdBlueStrong}18.34$\pm$1.66 & \cellcolor{opdBlueStrong}45.20$\pm$1.77 & \cellcolor{opdBlueStrong}\textbf{79.57$\pm$0.92} & \cellcolor{opdBlueStrong}78.93$\pm$0.01 & \cellcolor{opdBlueStrong}82.50$\pm$1.70 \\
 & \cellcolor{opdBlue}TA-OPD$_{+Ent.}$ & \cellcolor{opdBlue}54.10 & \cellcolor{opdBlue}21.67$\pm$5.00 & \cellcolor{opdBlue}\textbf{21.67$\pm$1.67} & \cellcolor{opdBlue}42.17$\pm$0.25 & \cellcolor{opdBlue}78.35$\pm$1.53 & \cellcolor{opdBlue}77.54$\pm$1.02 & \cellcolor{opdBlue}83.20$\pm$0.60 \\
\midrule
\multirow{6}{*}{\cellcolor{white}\shortstack[l]{DeepSeek-R1\\Distill-Qwen-14B\\$\rightarrow$\\Qwen2.5-3B}} & \cellcolor{opdGray}Base & \cellcolor{opdGray}29.98 & \cellcolor{opdGray}3.34$\pm$3.33 & \cellcolor{opdGray}1.67$\pm$1.66 & \cellcolor{opdGray}27.81$\pm$0.25 & \cellcolor{opdGray}64.63$\pm$1.22 & \cellcolor{opdGray}\textbf{26.16$\pm$0.27} & \cellcolor{opdGray}\textbf{56.31$\pm$0.20} \\
 & \cellcolor{opdGold}Pure OPD & \cellcolor{opdGold}28.76 & \cellcolor{opdGold}5.00$\pm$1.67 & \cellcolor{opdGold}1.67$\pm$1.66 & \cellcolor{opdGold}27.78$\pm$1.51 & \cellcolor{opdGold}62.20$\pm$1.21 & \cellcolor{opdGold}23.48$\pm$0.18 & \cellcolor{opdGold}52.50$\pm$0.90 \\
 & Entropy & 19.94 & 1.67$\pm$1.66 & 1.67$\pm$1.66 & 23.98$\pm$1.02 & 54.88$\pm$3.66 & 24.68$\pm$0.83 & 12.83$\pm$1.40 \\
 & TIP & 28.11 & 3.33$\pm$0.01 & 1.67$\pm$1.66 & 27.55$\pm$0.01 & 58.54$\pm$0.60 & \textbf{26.16$\pm$0.28} & 51.40$\pm$0.50 \\
 & \cellcolor{opdBlueStrong}TA-OPD & \cellcolor{opdBlueStrong}\textbf{30.62} & \cellcolor{opdBlueStrong}5.00$\pm$1.66 & \cellcolor{opdBlueStrong}1.67$\pm$1.66 & \cellcolor{opdBlueStrong}29.59$\pm$0.51 & \cellcolor{opdBlueStrong}\textbf{66.77$\pm$0.31} & \cellcolor{opdBlueStrong}25.42$\pm$1.20 & \cellcolor{opdBlueStrong}55.31$\pm$0.60 \\
 & \cellcolor{opdBlue}TA-OPD$_{+Ent.}$ & \cellcolor{opdBlue}30.28 & \cellcolor{opdBlue}\textbf{8.34$\pm$1.60} & \cellcolor{opdBlue}\textbf{3.33$\pm$0.01} & \cellcolor{opdBlue}\textbf{31.63$\pm$0.51} & \cellcolor{opdBlue}62.81$\pm$1.22 & \cellcolor{opdBlue}23.39$\pm$1.02 & \cellcolor{opdBlue}52.20$\pm$0.70 \\
\bottomrule
\end{tabular}
}
\end{table*}

%% file: tables/tab_budget_sweep_qwen3.tex
\begin{table*}[t]
\centering
\scriptsize
\setlength{\tabcolsep}{2.15pt}
\renewcommand{\arraystretch}{1.03}
\caption{Budget sweep for the two Qwen3-4B student settings.  
Benchmark entries report mean$\pm$std; 
Avg. is the arithmetic mean of the six benchmark means, 
excluding the $\pm$ terms. TA-OPD denotes Teachability-Aware OPD; 
TA-OPD$_{+Ent.}$ adds the entropy axis.}
\label{tab:budget-sweep-qwen3}
\resizebox{\textwidth}{!}{%
\begin{tabular}{l|c|lrrrrrrr}
\toprule
Setting & Budget & Selector & Avg. & AIME24 & AIME25 & GPQA-D. & HumanEval & IFEval & MATH-500 \\
\midrule
\multirow{16}{*}{\cellcolor{white}\shortstack[l]{Qwen3-8B\\(GRPO)$\rightarrow$\\Qwen3-4B}} & \multirow{4}{*}{\cellcolor{white}5\%} & Entropy & 56.95 & 31.66$\pm$1.66 & 26.66$\pm$0.01 & 47.47$\pm$0.15 & 76.83$\pm$1.22 & 73.47$\pm$0.64 & 85.60$\pm$0.20 \\
 &  & TIP & 56.20 & 21.65$\pm$8.35 & 24.98$\pm$1.68 & \textbf{48.23$\pm$0.25} & 78.66$\pm$0.01 & \textbf{76.80$\pm$0.28} & 86.90$\pm$0.50 \\
 &  & \cellcolor{opdBlueStrong}TA-OPD & \cellcolor{opdBlueStrong}57.35 & \cellcolor{opdBlueStrong}31.66$\pm$1.67 & \cellcolor{opdBlueStrong}26.67$\pm$0.01 & \cellcolor{opdBlueStrong}46.47$\pm$0.51 & \cellcolor{opdBlueStrong}78.97$\pm$0.92 & \cellcolor{opdBlueStrong}72.46$\pm$0.37 & \cellcolor{opdBlueStrong}\textbf{87.90$\pm$1.10} \\
 &  & \cellcolor{opdBlue}TA-OPD$_{+Ent.}$ & \cellcolor{opdBlue}\textbf{57.89} & \cellcolor{opdBlue}\textbf{33.34$\pm$6.67} & \cellcolor{opdBlue}\textbf{28.33$\pm$1.67} & \cellcolor{opdBlue}47.22$\pm$2.28 & \cellcolor{opdBlue}\textbf{79.58$\pm$0.92} & \cellcolor{opdBlue}72.37$\pm$0.28 & \cellcolor{opdBlue}86.50$\pm$0.30 \\
\addlinespace[0.15em]
 & \multirow{4}{*}{\cellcolor{white}10\%} & Entropy & 56.69 & 26.33$\pm$3.10 & 23.33$\pm$3.30 & \textbf{47.47$\pm$1.01} & 77.65$\pm$1.21 & \textbf{79.85$\pm$0.37} & 85.50$\pm$0.10 \\
 &  & TIP & 56.81 & 26.33$\pm$3.60 & 23.67$\pm$3.33 & 47.47$\pm$0.49 & \textbf{80.49$\pm$0.61} & 76.71$\pm$1.11 & 86.20$\pm$0.30 \\
 &  & \cellcolor{opdBlueStrong}TA-OPD & \cellcolor{opdBlueStrong}\textbf{56.87} & \cellcolor{opdBlueStrong}\textbf{30.00$\pm$1.40} & \cellcolor{opdBlueStrong}21.31$\pm$1.32 & \cellcolor{opdBlueStrong}\textbf{47.47$\pm$0.51} & \cellcolor{opdBlueStrong}79.88$\pm$0.60 & \cellcolor{opdBlueStrong}75.13$\pm$5.30 & \cellcolor{opdBlueStrong}\textbf{87.40$\pm$0.80} \\
 &  & \cellcolor{opdBlue}TA-OPD$_{+Ent.}$ & \cellcolor{opdBlue}56.52 & \cellcolor{opdBlue}\textbf{30.00$\pm$3.33} & \cellcolor{opdBlue}\textbf{25.00$\pm$1.67} & \cellcolor{opdBlue}45.70$\pm$2.78 & \cellcolor{opdBlue}78.05$\pm$1.22 & \cellcolor{opdBlue}73.38$\pm$2.22 & \cellcolor{opdBlue}87.00$\pm$0.40 \\
\addlinespace[0.15em]
 & \multirow{4}{*}{\cellcolor{white}30\%} & Entropy & 57.56 & 28.33$\pm$5.00 & 25.00$\pm$5.00 & 48.49$\pm$1.52 & 79.58$\pm$0.31 & 77.45$\pm$3.33 & 86.50$\pm$0.10 \\
 &  & TIP & \textbf{58.46} & \textbf{31.67$\pm$5.00} & 28.33$\pm$1.66 & 45.70$\pm$0.25 & \textbf{79.88$\pm$0.01} & 78.56$\pm$1.85 & \textbf{86.60$\pm$1.00} \\
 &  & \cellcolor{opdBlueStrong}TA-OPD & \cellcolor{opdBlueStrong}57.28 & \cellcolor{opdBlueStrong}28.34$\pm$1.67 & \cellcolor{opdBlueStrong}28.33$\pm$1.67 & \cellcolor{opdBlueStrong}\textbf{49.19$\pm$0.30} & \cellcolor{opdBlueStrong}77.75$\pm$0.31 & \cellcolor{opdBlueStrong}74.96$\pm$3.24 & \cellcolor{opdBlueStrong}85.10$\pm$0.50 \\
 &  & \cellcolor{opdBlue}TA-OPD$_{+Ent.}$ & \cellcolor{opdBlue}57.46 & \cellcolor{opdBlue}25.00$\pm$1.67 & \cellcolor{opdBlue}\textbf{28.34$\pm$1.67} & \cellcolor{opdBlue}47.98$\pm$3.03 & \cellcolor{opdBlue}78.97$\pm$0.92 & \cellcolor{opdBlue}\textbf{78.75$\pm$0.93} & \cellcolor{opdBlue}85.70$\pm$0.50 \\
\addlinespace[0.15em]
 & \multirow{4}{*}{\cellcolor{white}50\%} & Entropy & 53.90 & 20.00$\pm$3.33 & 20.00$\pm$3.33 & 43.68$\pm$1.76 & 78.35$\pm$0.91 & 75.69$\pm$4.34 & 85.70$\pm$1.50 \\
 &  & TIP & 55.74 & 23.33$\pm$3.33 & 24.99$\pm$1.66 & 44.69$\pm$0.75 & 78.96$\pm$0.91 & 76.15$\pm$2.40 & \textbf{86.33$\pm$0.33} \\
 &  & \cellcolor{opdBlueStrong}TA-OPD & \cellcolor{opdBlueStrong}\textbf{57.90} & \cellcolor{opdBlueStrong}\textbf{31.67$\pm$4.34} & \cellcolor{opdBlueStrong}23.34$\pm$0.01 & \cellcolor{opdBlueStrong}\textbf{47.73$\pm$0.76} & \cellcolor{opdBlueStrong}\textbf{78.97$\pm$0.92} & \cellcolor{opdBlueStrong}\textbf{79.67$\pm$1.11} & \cellcolor{opdBlueStrong}86.00$\pm$1.80 \\
 &  & \cellcolor{opdBlue}TA-OPD$_{+Ent.}$ & \cellcolor{opdBlue}57.25 & \cellcolor{opdBlue}23.33$\pm$0.01 & \cellcolor{opdBlue}\textbf{30.00$\pm$3.33} & \cellcolor{opdBlue}46.97$\pm$1.01 & \cellcolor{opdBlue}78.05$\pm$0.61 & \cellcolor{opdBlue}78.93$\pm$0.19 & \cellcolor{opdBlue}86.20$\pm$0.40 \\
\midrule
\multirow{16}{*}{\cellcolor{white}\shortstack[l]{Qwen3-14B\\$\rightarrow$\\Qwen3-4B}} & \multirow{4}{*}{\cellcolor{white}5\%} & Entropy & 53.56 & \textbf{23.34$\pm$3.34} & 16.67$\pm$0.01 & 43.94$\pm$2.02 & 77.75$\pm$0.92 & 75.97$\pm$5.36 & \textbf{83.70$\pm$0.30} \\
 &  & TIP & 53.83 & 23.33$\pm$0.01 & 16.67$\pm$0.01 & 44.70$\pm$1.77 & \textbf{79.88$\pm$0.61} & 75.50$\pm$2.49 & 82.90$\pm$0.70 \\
 &  & \cellcolor{opdBlueStrong}TA-OPD & \cellcolor{opdBlueStrong}53.58 & \cellcolor{opdBlueStrong}18.34$\pm$1.66 & \cellcolor{opdBlueStrong}\textbf{18.34$\pm$1.66} & \cellcolor{opdBlueStrong}45.70$\pm$1.27 & \cellcolor{opdBlueStrong}77.75$\pm$0.30 & \cellcolor{opdBlueStrong}\textbf{79.66$\pm$0.93} & \cellcolor{opdBlueStrong}81.70$\pm$1.30 \\
 &  & \cellcolor{opdBlue}TA-OPD$_{+Ent.}$ & \cellcolor{opdBlue}\textbf{54.47} & \cellcolor{opdBlue}21.66$\pm$1.66 & \cellcolor{opdBlue}16.67$\pm$0.01 & \cellcolor{opdBlue}\textbf{48.23$\pm$2.27} & \cellcolor{opdBlue}77.75$\pm$0.91 & \cellcolor{opdBlue}79.02$\pm$0.65 & \cellcolor{opdBlue}83.50$\pm$0.70 \\
\addlinespace[0.15em]
 & \multirow{4}{*}{\cellcolor{white}10\%} & Entropy & 53.39 & 21.67$\pm$5.00 & 15.00$\pm$1.67 & 46.46$\pm$0.01 & 79.27$\pm$0.61 & 73.94$\pm$4.06 & \textbf{84.00$\pm$0.80} \\
 &  & TIP & 53.62 & 20.00$\pm$0.01 & 16.66$\pm$3.33 & \textbf{47.98$\pm$2.53} & 78.05$\pm$0.61 & 76.62$\pm$1.57 & 82.40$\pm$0.01 \\
 &  & \cellcolor{opdBlueStrong}TA-OPD & \cellcolor{opdBlueStrong}\textbf{54.65} & \cellcolor{opdBlueStrong}\textbf{23.34$\pm$3.34} & \cellcolor{opdBlueStrong}18.34$\pm$1.66 & \cellcolor{opdBlueStrong}45.20$\pm$1.77 & \cellcolor{opdBlueStrong}\textbf{79.57$\pm$0.92} & \cellcolor{opdBlueStrong}\textbf{78.93$\pm$0.01} & \cellcolor{opdBlueStrong}82.50$\pm$1.70 \\
 &  & \cellcolor{opdBlue}TA-OPD$_{+Ent.}$ & \cellcolor{opdBlue}54.10 & \cellcolor{opdBlue}21.67$\pm$5.00 & \cellcolor{opdBlue}\textbf{21.67$\pm$1.67} & \cellcolor{opdBlue}42.17$\pm$0.25 & \cellcolor{opdBlue}78.35$\pm$1.53 & \cellcolor{opdBlue}77.54$\pm$1.02 & \cellcolor{opdBlue}83.20$\pm$0.60 \\
\addlinespace[0.15em]
 & \multirow{4}{*}{\cellcolor{white}30\%} & Entropy & 53.62 & 20.00$\pm$3.33 & \textbf{21.66$\pm$1.66} & \textbf{46.97$\pm$0.01} & 78.97$\pm$0.92 & 70.52$\pm$0.46 & \textbf{83.60$\pm$0.20} \\
 &  & TIP & \textbf{53.93} & \textbf{26.67$\pm$0.01} & \textbf{21.66$\pm$5.00} & 43.18$\pm$1.26 & 79.57$\pm$0.30 & 69.97$\pm$0.09 & 82.50$\pm$0.50 \\
 &  & \cellcolor{opdBlueStrong}TA-OPD & \cellcolor{opdBlueStrong}52.81 & \cellcolor{opdBlueStrong}21.66$\pm$8.33 & \cellcolor{opdBlueStrong}15.00$\pm$1.67 & \cellcolor{opdBlueStrong}43.18$\pm$0.76 & \cellcolor{opdBlueStrong}79.57$\pm$0.30 & \cellcolor{opdBlueStrong}\textbf{74.77$\pm$3.42} & \cellcolor{opdBlueStrong}82.70$\pm$0.10 \\
 &  & \cellcolor{opdBlue}TA-OPD$_{+Ent.}$ & \cellcolor{opdBlue}53.15 & \cellcolor{opdBlue}20.00$\pm$3.33 & \cellcolor{opdBlue}18.34$\pm$1.66 & \cellcolor{opdBlue}45.70$\pm$2.77 & \cellcolor{opdBlue}\textbf{79.88$\pm$0.01} & \cellcolor{opdBlue}72.45$\pm$1.30 & \cellcolor{opdBlue}82.50$\pm$0.10 \\
\addlinespace[0.15em]
 & \multirow{4}{*}{\cellcolor{white}50\%} & Entropy & 52.76 & 18.34$\pm$1.66 & 18.33$\pm$1.67 & 44.95$\pm$0.51 & 78.66$\pm$0.01 & 73.38$\pm$2.59 & 82.90$\pm$0.50 \\
 &  & TIP & 52.12 & \textbf{21.66$\pm$1.66} & 13.33$\pm$0.01 & \textbf{45.95$\pm$1.52} & 78.05$\pm$0.01 & 70.34$\pm$0.27 & 83.40$\pm$1.40 \\
 &  & \cellcolor{opdBlueStrong}TA-OPD & \cellcolor{opdBlueStrong}52.16 & \cellcolor{opdBlueStrong}13.33$\pm$0.01 & \cellcolor{opdBlueStrong}\textbf{18.34$\pm$1.66} & \cellcolor{opdBlueStrong}45.20$\pm$0.25 & \cellcolor{opdBlueStrong}77.75$\pm$0.31 & \cellcolor{opdBlueStrong}73.75$\pm$3.51 & \cellcolor{opdBlueStrong}\textbf{84.60$\pm$0.20} \\
 &  & \cellcolor{opdBlue}TA-OPD$_{+Ent.}$ & \cellcolor{opdBlue}\textbf{52.95} & \cellcolor{opdBlue}18.34$\pm$1.66 & \cellcolor{opdBlue}16.66$\pm$3.34 & \cellcolor{opdBlue}44.19$\pm$2.78 & \cellcolor{opdBlue}\textbf{78.97$\pm$0.91} & \cellcolor{opdBlue}\textbf{76.43$\pm$1.94} & \cellcolor{opdBlue}83.10$\pm$0.10 \\
\bottomrule
\end{tabular}
}
\vspace{-1em}
\end{table*}

%% file: tables/tab_section3_dataset_manifest.tex
\begin{table*}[t]
\centering
\small
\caption{Section~3 diagnostic datasets.  The short diagnostic run is retained only as a boundary check.}
\label{tab:section3-dataset-manifest}
\resizebox{0.55\textwidth}{!}{%
\begin{tabular}{lrrl}
\hline
Dataset & Contexts & Tokens & Use \\
\hline
4B $\rightarrow$ 1.7B held-out 300 & 300 & 57,600 & main \\
4B $\rightarrow$ 1.7B GSM8K-COT 300 & 300 & 57,600 & main \\
8B $\rightarrow$ 1.7B 300 & 300 & 57,600 & main \\
14B $\rightarrow$ 1.7B 300 K=8 & 300 & 57,600 & main \\
14B $\rightarrow$ 1.7B 300 & 300 & 57,600 & main \\
14B $\rightarrow$ 1.7B 300 K=32 & 300 & 57,600 & main \\
8B $\rightarrow$ 1.7B short diagnostic & 8 & 1,024 & boundary only \\
\hline
\end{tabular}%
}
\end{table*}

%% file: tables/tab_support_proxy_audit.tex
\begin{table*}[t]
\centering
\small
\caption{Support proxy audit inside the low-entropy with high-divergence (Q3) region.  Different support definitions produce positive high--low gain gaps.}
\label{tab:support-proxy-audit}
\resizebox{0.65\textwidth}{!}{%
\begin{tabular}{llrrr}
\hline
Setting & Proxy & Tokens & Gap & 95\% interval \\
\hline
14B $\rightarrow$ 1.7B, K=16 & $C_t$ mass & 3342 & +0.086 & [+0.064, +0.104] \\
14B $\rightarrow$ 1.7B, K=16 & CBC & 3342 & +0.105 & [+0.078, +0.125] \\
14B $\rightarrow$ 1.7B, K=16 & Jaccard & 3342 & +0.038 & [+0.020, +0.055] \\
14B $\rightarrow$ 1.7B, K=16 & overlap & 3342 & +0.038 & [+0.020, +0.055] \\
14B $\rightarrow$ 1.7B, K=32 & $C_t$ mass & 3342 & +0.090 & [+0.074, +0.106] \\
14B $\rightarrow$ 1.7B, K=32 & CBC & 3342 & +0.105 & [+0.078, +0.125] \\
14B $\rightarrow$ 1.7B, K=32 & Jaccard & 3342 & +0.052 & [+0.032, +0.069] \\
14B $\rightarrow$ 1.7B, K=32 & overlap & 3342 & +0.052 & [+0.032, +0.069] \\
\hline
\end{tabular}%
}
\end{table*}

%% file: tables/tab_within_q3_live.tex
\begin{table*}[t]
\centering
\small
\caption{Low-entropy with high-divergence live intervention. Even inside this TIP region, high teachability is beneficial while low teachability is harmful.}
\label{tab:within-q3-live}
\resizebox{0.75\textwidth}{!}{%
\begin{tabular}{lrrrrrr}
\toprule
Q3-region mask & Keep \% & Boot. gain & Match. gain & $D^L$ & $D^I$ & $C$ \\
\midrule
Q3+$D^L$ high & 3.26 & 0.015 $\pm$ 0.010 & 0.010 & 0.072 & 0.005 & 0.972 \\
Q3+$D^L$ low & 3.24 & -0.010 $\pm$ 0.015 & -0.006 & 0.015 & 0.011 & 0.933 \\
Q3+$D^I$ high & 3.23 & 0.006 $\pm$ 0.003 & -0.003 & 0.053 & 0.012 & 0.919 \\
\bottomrule
\end{tabular}
}
\end{table*}

%% file: tables/tab_matched_topn.tex
\begin{table*}[t]
\centering
\small
\caption{Exact top-$N$ matched fixed-context analysis at $K=16$.
Useful as a confound check: all selectors retain exactly the same number of tokens.
}
\label{tab:matched-topn}
\resizebox{0.8\textwidth}{!}{%
\begin{tabular}{lrrrrrrrr}
\toprule
Selector & Ratio & Gain & 95\% interval & Gain/keep & Q3 frac. & $H$ & $D$ & $C$ \\
\midrule
$D^L$ & 3 & 0.130 & [0.123, 0.138] & 4.34 & 7.4 & 0.31 & 1.00 & 0.99 \\
$D^L$ & 5 & 0.135 & [0.128, 0.143] & 2.70 & 5.5 & 0.37 & 0.94 & 0.97 \\
KL & 3 & 0.120 & [0.112, 0.129] & 4.00 & 6.1 & 0.40 & 1.00 & 0.66 \\
KL & 5 & 0.119 & [0.113, 0.126] & 2.38 & 5.4 & 0.41 & 1.00 & 0.65 \\
TIP-soft & 3 & 0.093 & [0.075, 0.110] & 3.11 & 2.6 & 0.70 & 0.69 & 0.59 \\
TIP-soft & 5 & 0.094 & [0.080, 0.107] & 1.88 & 2.6 & 0.69 & 0.69 & 0.61 \\
Entropy & 3 & 0.076 & [0.055, 0.095] & 2.53 & 0.0 & 1.00 & 0.42 & 0.54 \\
Entropy & 5 & 0.066 & [0.050, 0.082] & 1.32 & 0.0 & 1.00 & 0.42 & 0.56 \\
Random & 3 & -0.130 & [-0.200, -0.072] & -4.35 & 4.8 & 0.18 & 0.13 & 0.93 \\
Random & 5 & -0.105 & [-0.149, -0.057] & -2.09 & 4.8 & 0.19 & 0.13 & 0.93 \\
$D^I$ & 3 & 0.105 & [0.096, 0.113] & 3.50 & 1.8 & 0.64 & 0.87 & 0.03 \\
$D^I$ & 5 & 0.111 & [0.104, 0.117] & 2.22 & 1.6 & 0.66 & 0.72 & 0.11 \\
\bottomrule
\end{tabular}
}
\vspace{0.25em}
\end{table*}

%% file: tables/tab_decomposition_regression.tex
\begin{table*}[t]
\centering
\small
\caption{Prompt-cluster bootstrap regression decomposition. 
Learnable disagreement has a consistently larger standardized coefficient than incompatible disagreement.
$\Delta R^2_{\rm decomp}$ is reported in $10^{-3}$ units relative to the entropy+divergence baseline.
}
\label{tab:decomposition-regression}
\resizebox{0.65\textwidth}{!}{%
\begin{tabular}{rrrrrrr}
\toprule
$K$ & Q3 tokens & $\beta_{L}$ & $\beta_{I}$ & Gap & 95\% interval & $\Delta R^2_{\rm decomp}$ \\
\midrule
8 & 2393 & 0.086 & 0.045 & 0.041 & [0.039, 0.044] & 0.897 \\
16 & 2393 & 0.087 & 0.043 & 0.044 & [0.041, 0.047] & 0.790 \\
32 & 2393 & 0.087 & 0.043 & 0.044 & [0.041, 0.047] & 0.743 \\
\bottomrule
\end{tabular}
}
\vspace{0.25em}
\end{table*}

%% file: tables/tab_budget_curve.tex
\begin{table*}[t]
\centering
\small
\caption{Fixed-context budget sweep for TA-OPD. The teachability selector peaks around 3--5\% effective target-token budget and saturates at 10\%.}
\label{tab:budget-sweep}
\resizebox{0.65\textwidth}{!}{%
\begin{tabular}{lrrrrr}
\toprule
Selector & Nominal \% & Actual \% & Bootstrap gain & Matching gain & Seeds \\
\midrule
TA-OPD & 1 & 1.89 & 0.034 $\pm$ 0.016 & 0.017 $\pm$ 0.015 & 3 \\
\rowcolor{opdBlueStrong}
TA-OPD & 3 & 3.35 & 0.047 $\pm$ 0.006 & 0.030 $\pm$ 0.003 & 3 \\
\rowcolor{opdBlueStrong}
TA-OPD & 5 & 5.14 & 0.049 $\pm$ 0.002 & 0.027 $\pm$ 0.003 & 3 \\
TA-OPD & 10 & 10.16 & 0.035 $\pm$ 0.013 & 0.022 $\pm$ 0.007 & 3 \\
\rowcolor{opdGray}
Q3+high-$C$ & 1 & 1.93 & 0.009 $\pm$ 0.004 & 0.002 $\pm$ 0.002 & 2 \\
\rowcolor{opdGray}
Q3+high-$C$ & 3 & 3.20 & 0.016 $\pm$ 0.001 & 0.007 $\pm$ 0.006 & 3 \\
\rowcolor{opdGray}
Q3+high-$C$ & 5 & 3.93 & 0.009 $\pm$ 0.015 & 0.002 $\pm$ 0.009 & 3 \\
Q3+low-$C$ & 3 & 3.23 & 0.010 $\pm$ 0.003 & 0.006 $\pm$ 0.006 & 3 \\
$D^I$ & 3 & 3.52 & 0.025 $\pm$ 0.000 & 0.014 $\pm$ 0.005 & 3 \\
\bottomrule
\end{tabular}
}
\end{table*}

%% file: tables/tab_budget_ratio_qwen3.tex
\begin{table*}[t]
\centering
\small
\setlength{\tabcolsep}{3.8pt}
\renewcommand{\arraystretch}{1.08}
\caption{Macro-average budget sweep on the two Qwen3-4B student settings.  Scores average the six downstream benchmark means in Table~\ref{tab:budget-sweep-qwen3}; the full per-benchmark results are kept in the main table.  TA-OPD denotes Teachability-Aware OPD, and TA-OPD$_{+Ent.}$ adds the entropy axis.}
\label{tab:budget-ratio-qwen3}
\resizebox{0.55\textwidth}{!}{%
\begin{tabular}{llrrrr}
\toprule
Setting & Budget & Entropy & TIP & TA-OPD & TA-OPD$_{+Ent.}$ \\
\midrule
\multirow{4}{*}{\shortstack[l]{Qwen3-8B\\(GRPO)$\rightarrow$4B}}
& 5\%  & 56.95 & 56.20 & 57.35 & \cellcolor{opdBlueStrong}\textbf{57.89} \\
& 10\% & 56.69 & 56.81 & \cellcolor{opdBlue}56.87 & 56.52 \\
& 30\% & 57.56 & \cellcolor{opdGray}\textbf{58.46} & 57.28 & 57.46 \\
& 50\% & 53.90 & 55.74 & \cellcolor{opdBlueStrong}\textbf{57.90} & 57.25 \\
\midrule
\multirow{4}{*}{\shortstack[l]{Qwen3-14B\\$\rightarrow$4B}}
& 5\%  & 53.56 & 53.83 & 53.58 & \cellcolor{opdBlueStrong}\textbf{54.47} \\
& 10\% & 53.39 & 53.62 & \cellcolor{opdBlueStrong}\textbf{54.65} & 54.10 \\
& 30\% & 53.62 & \cellcolor{opdGray}\textbf{53.93} & 52.81 & 53.15 \\
& 50\% & 52.76 & 52.12 & 52.16 & \cellcolor{opdBlue}\textbf{52.95} \\
\bottomrule
\end{tabular}
}
\end{table*}

%% file: tables/tab_downstream_selector_ablation.tex
\begin{table*}[t]
\centering
\scriptsize
\setlength{\tabcolsep}{3.0pt}
\renewcommand{\arraystretch}{1.06}
\caption{Downstream selector ablation for Qwen3-8B-GRPO $\rightarrow$ Qwen3-4B at a 10\% supervised-token budget. Entries report mean$\pm$std over five evaluation seeds; Avg. averages the six benchmark means. TA-OPD is not reducible to raw KL, compatibility mass $C$, or the low-entropy with high-divergence Q3 region.}
\label{tab:downstream-selector-ablation}
\resizebox{\textwidth}{!}{%
\begin{tabular}{lrrrrrrr}
\toprule
Selector & Avg. & AIME24 & AIME25 & GPQA-D. & HumanEval & IFEval & MATH-500 \\
\midrule
\rowcolor{opdGold}
Raw KL & 53.76 & 21.66$\pm$1.66 & 20.00$\pm$0.01 & 44.95$\pm$1.52 & 78.97$\pm$0.30 & 73.38$\pm$2.22 & 83.60$\pm$0.40 \\
\rowcolor{opdGray}
$C$-only & 54.19 & 21.66$\pm$1.66 & 18.33$\pm$8.33 & 43.94$\pm$3.03 & 77.75$\pm$0.91 & \textbf{80.13$\pm$0.83} & 83.30$\pm$0.30 \\
\rowcolor{opdGray}
Q3-only & 52.46 & 21.66$\pm$1.66 & 15.00$\pm$1.67 & 44.70$\pm$1.27 & \textbf{79.57$\pm$0.91} & 70.52$\pm$0.28 & 83.30$\pm$0.10 \\
\rowcolor{opdGray}
Q3+high-$C$ & 53.32 & 21.67$\pm$5.00 & 16.66$\pm$3.33 & 44.69$\pm$3.29 & 78.97$\pm$0.30 & 75.05$\pm$5.36 & 82.90$\pm$1.10 \\
Entropy & 53.39 & 21.67$\pm$5.00 & 15.00$\pm$1.67 & 46.46$\pm$0.01 & 79.27$\pm$0.61 & 73.94$\pm$4.06 & \textbf{84.00$\pm$0.80} \\
TIP & 53.62 & 20.00$\pm$0.01 & 16.66$\pm$3.33 & \textbf{47.98$\pm$2.53} & 78.05$\pm$0.61 & 76.62$\pm$1.57 & 82.40$\pm$0.01 \\
\rowcolor{opdBlue}
TA-OPD$_{+Ent.}$ & 54.10 & 21.67$\pm$5.00 & \textbf{21.67$\pm$1.67} & 42.17$\pm$0.25 & 78.35$\pm$1.53 & 77.54$\pm$1.02 & 83.20$\pm$0.60 \\
\rowcolor{opdBlueStrong}
TA-OPD & \textbf{54.65} & \textbf{23.34$\pm$3.34} & 18.34$\pm$1.66 & 45.20$\pm$1.77 & \textbf{79.57$\pm$0.92} & 78.93$\pm$0.01 & 82.50$\pm$1.70 \\
\bottomrule
\end{tabular}
}
\end{table*}

%% file: tables/tab_math_category_deltas.tex
\begin{table*}[t]
\centering
\small
\caption{MATH-hard category deltas relative to the base model.
Values are percentage-point changes. This table is best treated as boundary analysis rather than the central claim.
}
\label{tab:math-category-deltas}
\resizebox{0.7\textwidth}{!}{%
\begin{tabular}{lrrrrrrr}
\toprule
Model & Alg. & C\&P & Geo. & IntAlg. & Num. & PreAlg. & PreCalc. \\
\midrule
$D^L$ & +1.6 & 0.0 & +3.0 & -1.4 & -3.9 & -0.5 & +1.5 \\
TIP & +2.9 & +0.8 & +2.3 & +3.2 & -7.1 & +3.1 & +0.7 \\
$D^I$ & +1.0 & +4.9 & +0.8 & +1.8 & -3.9 & +2.6 & 0.0 \\
Random & +2.0 & -0.8 & -0.8 & +0.4 & -3.9 & +2.1 & 0.0 \\
Entropy+$D^L$ & 0.0 & +0.8 & +3.8 & +1.4 & -5.2 & 0.0 & +0.7 \\
Entropy & 0.0 & 0.0 & +2.3 & -2.1 & -2.0 & +0.5 & -0.7 \\
\bottomrule
\end{tabular}
}
\vspace{0.25em}
\end{table*}